\documentclass[a4paper,fleqn]{cas-dc}

\usepackage[numbers]{natbib}
\usepackage{multirow,booktabs}
\usepackage{amsmath}
\usepackage{algorithm}
\usepackage{algorithmic}
\usepackage{color}
\usepackage[table,HTML]{xcolor}
\usepackage{enumitem}
\usepackage{pifont}

\usepackage{ragged2e} 
\usepackage{threeparttable}
\usepackage{multirow} 

\def\tsc#1{\csdef{#1}{\textsc{\lowercase{#1}}\xspace}}
\tsc{WGM}
\tsc{QE}
\tsc{EP}
\tsc{PMS}
\tsc{BEC}
\tsc{DE}

\makeatletter
\renewcommand{\section}{\@startsection{section}{1}{\z@}%
  {-3.5ex \@plus -1ex \@minus -.2ex}
  {2.3ex \@plus.2ex}
  {\normalfont\bfseries\normalsize}} 
\renewcommand{\subsection}{\@startsection{subsection}{2}{\z@}%
  {-3.25ex \@plus -1ex \@minus -.2ex}%
  {1.5ex \@plus .2ex}%
  {\normalfont\bfseries\normalsize}}
\makeatother

\begin{document}
\let\WriteBookmarks\relax
\def\floatpagepagefraction{1}
\def\textpagefraction{.001}

\title [mode = title]{$\Psi$-Map: Panoptic Surface Integrated Mapping Enables Real2Sim Transfer}                      



\author[1]{Xuan Yu}
\author[1]{Yuxuan Xie}
\author[1]{Changjian Jiang}
\author[1]{Shichao Zhai}
\author[1]{Rong Xiong}
\author[1]{Yu Zhang}
\author[1]{Yue Wang}\cormark[1]

\cortext[cor1]{Corresponding author}

\ead{ywang24@zju.edu.cn}

\affiliation[1]{%
  organization={Department of Control Science and Engineering, Zhejiang University},
  addressline={},
  city={Hangzhou},
  state={Zhejiang},
  postcode={},
  country={P.~R.~China}
}

\begin{abstract}
Open-vocabulary panoptic reconstruction is essential for advanced robotics perception and simulation. However, existing methods based on 3D Gaussian Splatting (3DGS) often struggle to simultaneously achieve geometric accuracy, coherent panoptic understanding, and real-time inference frequency in large-scale scenes. In this paper, we propose a comprehensive framework that integrates geometric reinforcement, end-to-end panoptic learning, and efficient rendering. First, to ensure physical realism in large-scale environments, we leverage LiDAR data to construct plane-constrained multimodal Gaussian Mixture Models (GMMs) and employ 2D Gaussian surfels as the map representation, enabling high-precision surface alignment and continuous geometric supervision. Building upon this, to overcome the error accumulation and cumbersome cross-frame association inherent in traditional multi-stage panoptic segmentation pipelines, we design a query-guided end-to-end learning architecture. By utilizing a local cross-attention mechanism within the view frustum, the system lifts 2D mask features directly into 3D space, achieving globally consistent panoptic understanding. Finally, addressing the computational bottlenecks caused by high-dimensional semantic features, we introduce Precise Tile Intersection and a Top-K Hard Selection strategy to optimize the rendering pipeline. Experimental results demonstrate that our system achieves superior geometric and panoptic reconstruction quality in large-scale scenes while maintaining an inference rate exceeding 50 FPS, meeting the real-time requirements of robotic control loops.
\end{abstract}




\maketitle

\section{Introduction}

In the contemporary landscape of robotics, learning from simulation has emerged as the mainstream paradigm for developing robust and high-performance policies~\cite{tao2024maniskill3, mittal2025isaac, zhang2024uni-navid}. However, the deployment of these policies in physical environments is often hindered by the Sim2Real Gap, a fundamental discrepancy between synthetic training data and real-world dynamics. To facilitate successful Sim2Real transfer, it is imperative to construct a high-fidelity Real2Sim mapping, enabling a digital twin pipeline capable of migrating physical environment characteristics into simulation to support environment-specific policy adaptation.

Effective Real2Sim transfer demands the resolution of three core challenges. First, Geometric Integrity is paramount. The system must produce high-quality surface meshes rather than sparse point clouds to support collision detection and contact dynamics within physics engines. Second, Panoptic Understanding is required to enable object-level reasoning. The reconstruction must move beyond holistic maps to identify discrete, manipulatable entities for complex interactions. Finally, High-efficiency Rendering is essential. The system must provide photorealistic textures aligned with the geometry while meeting the high-frequency frame-rate requirements of robotic simulation loops.

Current Real2Sim research primarily focuses on tabletop manipulation within small-scale indoor environments~\cite{li2024evaluating, wu2025rlgsbridge, qureshi2025splatsim}. These methods rely heavily on monocular techniques, such as monocular depth estimation, single-view semantics, and generative priors, which are difficult to scale or validate in expansive, complex scenes~\cite{li2025rose, han2025re3sim}. Conversely, existing large-scale reconstruction frameworks often prioritize pure visual fidelity over metric accuracy, utilizing vision-only pipelines that lack the geometric precision required for physical simulation and often ignore semantic integration~\cite{xia2024video2game, wu2023objectsdf++, wang2025vggt}. While individual modules for depth, semantics, or rendering exist, the strategic integration of these components into a unified system solution that avoids cumulative error and supports downstream policy verification remains a critical gap for deployment of the learning.

\begin{figure*}
    \centering
    \includegraphics[width=\linewidth]{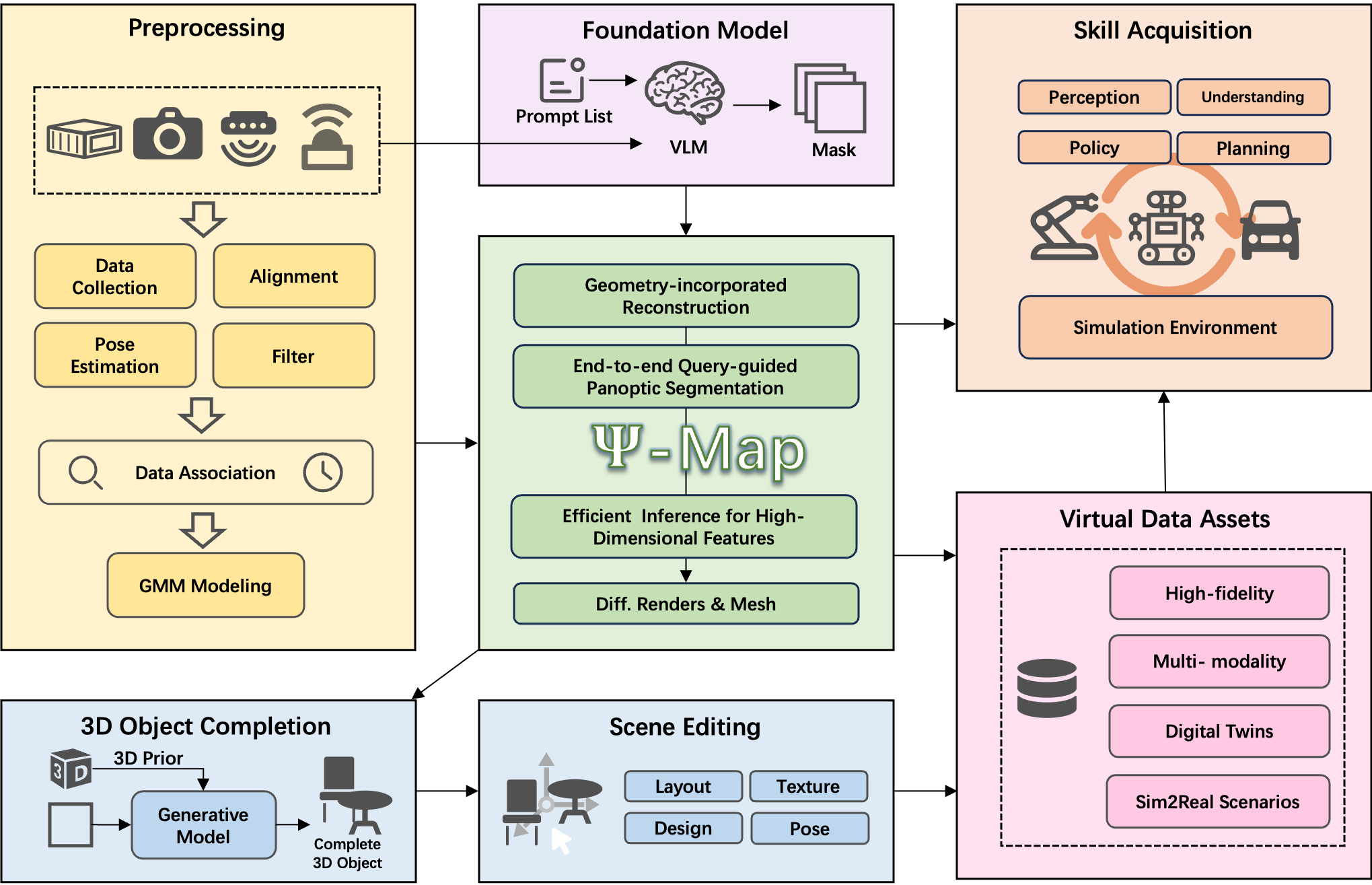}
    \caption{$\Psi$-Map provides a high-performance bridge from multi-modal sensor inputs to robotic applications. By fusing LiDAR-RGB-D data with Vision-Language Foundation Models, our framework achieves real-time, 3D-consistent panoptic reconstruction. This representation serves as a powerful engine for high-fidelity scene editing and virtual asset generation, ultimately empowering complex downstream skills such as generative 3D object completion and robust robotic navigation in wild environments.}
    \label{fig:token}
\end{figure*}

To address these limitations, we propose $\Psi$-Map, a comprehensive panoptic reconstruction system engineered to enable seamless Real2Sim transfer Fig.~\ref{fig:token}. The core of $\Psi$-Map is an end-to-end semantic reconstruction and rendering framework that fuses RGB and depth data to satisfy both metric geometric and panoptic semantic requirements. Building upon this, we implement an accelerated Gaussian and surface rendering architecture that ensures real-time performance for robot simulation. Finally, we integrate a downstream object generation model capable of recovering complete 3D object geometries from partially observed segments, directly serving robotic manipulation tasks. By utilizing three low-coupling yet highly efficient modules, $\Psi$-Map fulfills the entire spectrum of Real2Sim requirements. Our experiments demonstrate the excellence of each module, culminating in a robot navigation task in unseen environments that validates the system's ability to facilitate policy deployment. We anticipate that $\Psi$-Map will significantly advance the deployment capabilities of robotic learning in real-world applications. Our core contributions are as follows:
\begin{itemize}
    \item We present a low-coupling system solution that bridges the real sensors data to actionable digital twins. This design minimizes cumulative errors across modules, providing a robust foundation for scene-specific policy transfer.
    \item We fuse depth-based GMM initialization with query-guided panoptic lifting, which ensures both physical surface and semantics. This allows for precise collision dynamics and semantic interaction within the simulation.
    \item We introduce an accelerated semantic rendering framework achieving over 50 FPS and a generative model for 3D object completion, providing the fast visual feedback and geometric wholeness necessary for physical interaction.
    \item We verify the pipeline through robot navigation experiments in previously unseen environments. Results show that fine-tuning policies within the $\Psi$-Map generated simulation effectively enhances the performance.
\end{itemize}

\section{Related Work}

\subsection{3D Gaussian Splatting and Geometric Modeling}
3D Gaussian Splatting (3DGS) \cite{feature3dgs} has revolutionized radiance field representation by utilizing an explicit point-based structure and a tile-based rasterization pipeline to achieve exceptional rendering quality. While vanilla 3DGS performs remarkably in small-scale environments, it often suffers from geometric collapse, holes, or excessive "floaters" in large-scale or unbounded scenes due to a lack of strong structural regularization during the initialization process, which typically relies on sparse Structure-from-Motion (SfM) point clouds.

To mitigate these robustness issues, various geometric reinforcement strategies have been proposed. Mip-Splatting \cite{yu2024mip} introduced anti-aliasing filters to enhance multi-scale stability, while SuGaR \cite{guedon2024sugar} and 2DGS \cite{2dgs} optimized surface reconstruction by constraining 3D ellipsoids into 2D Gaussian surfels with defined normals. These advancements significantly improve the representation of thin-walled structures and planar regions, providing a more reliable physical foundation for downstream robotic interaction tasks.

Furthermore, for large-scale reconstruction, research has explored incorporating depth map supervision \cite{chen2024pgsr} or multi-view geometric consistency constraints. However, vision-based depth estimation remains uncertain in texture-less regions or areas with dramatic lighting changes. Unlike existing methods purely dependent on visual feedback, we propose leveraging point cloud priors to construct plane-constrained multimodal Gaussian Mixture Models (GMMs) \cite{gmm, gmm2}. This approach provides a reliable scale reference during initialization and ensures, through a GMM-guided optimization process, that surfels adhere to the underlying geometric structure, maintaining high physical realism in challenging environments.

\subsection{3D Panoptic Scene Understanding}
Developing semantically-aware 3D representations is a fundamental requirement for robots to perceive their environment and execute complex interaction tasks. Early efforts focused on embedding open-vocabulary features, such as CLIP \cite{clip}, into radiance fields like LERF \cite{kerr2023lerf} and LangSplat \cite{langsplat} to enable zero-shot language-driven object localization. These works demonstrated the potential of aligning vision-language features with 3D space but remained limited in distinguishing between multiple instances with identical semantic attributes.

With the rising demand for panoptic segmentation, methods like Gaussian Grouping \cite{gaussiangrouping} attemptes to lift instance labels into 3D space via multi-stage pipelines. However, these paradigms rely heavily on pseudo-labels from 2D models \cite{sam} and require cumbersome, hand-designed association algorithms. Since 2D predictions often exhibit flickering and category instability across views, these manual association logics are prone to error accumulation and perform poorly under dynamic occlusions.

Inspired by end-to-end panoptic architectures \cite{mask2former}, we design a query-guided learning mechanism. By applying local cross-attention within the view frustum, the system models instance attributes directly within the 3D Gaussian feature space, effectively bypassing the need for explicit cross-frame association. This design not only streamlines the pipeline but also allows the system to rectify noise in 2D labels during global optimization. Additionally, we introduce label blending and warping techniques to further enhance multi-view segmentation consistency.

\subsection{Efficient Neural Rendering and Robotic Simulation}
While 3DGS exhibits a natural advantage in RGB rendering, it faces significant performance bottlenecks when integrating high-dimensional semantic or instance features. Existing semantic field solutions often struggle to maintain inference frequencies above 10 FPS when processing feature maps with hundreds of dimensions. This latency primarily stems from frequent memory bandwidth swaps and high computational complexity during feature accumulation, failing to meet the stringent real-time requirements of robotic control loops.

To address this challenge, some studies have explored feature quantization, pruning, or low-rank decomposition \cite{lee2024compact}. Although these solutions reduce memory overhead, they often sacrifice segmentation accuracy and do not fundamentally optimize the computational flow of the rasterization pipeline. Balancing rendering resolution with high-dimensional features in large-scale scenes remains a key technical hurdle for practical 3D Gaussian applications in simulation and interaction.

\begin{figure*}[t]
    \centering
    \includegraphics[width=\linewidth,keepaspectratio]{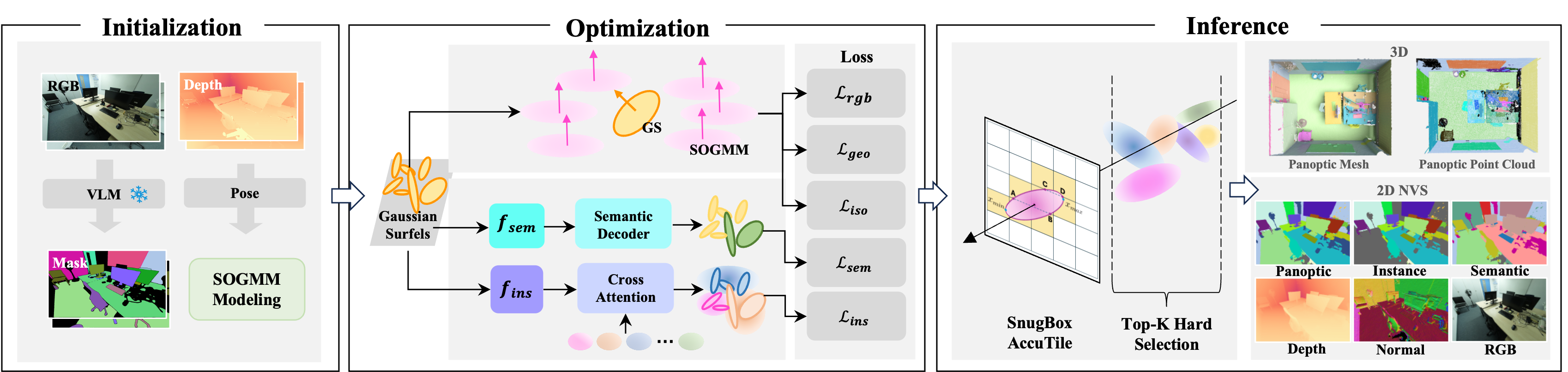}\\
    \caption{Overview of the $\Psi$-Map Framework.
The Input consists of multi-view RGB-D frames. During Geometric Reinforcement, the point cloud is modeled as a plane-constrained SOGMM to provide continuous structural supervision for the 2D Gaussian Surfels. In the Panoptic Learning stage, 2D mask features are lifted into 3D via a query-guided end-to-end architecture, where instance tokens are refined through local cross-attention within the view frustum to ensure global identity consistency. For Inference, our highly optimized rendering pipeline employs Precise Tile Intersection and Top-$K$ Hard Selection. By leveraging the geometric sparsity and surface-aligned nature of 2DGS, $\Psi$-Map significantly reduces the computational overhead of high-dimensional feature accumulation, enabling real-time, high-fidelity multi-modal rendering.}
    \label{fig:first_fig}
\end{figure*}

\section{Geometric Representation and Reconstruction}

\subsection{Preliminaries and 2D Surfel Representation}
In large-scale scene reconstruction, traditional 3DGS often suffers from geometric collapse due to the lack of structural constraints on its anisotropic 3D ellipsoids. To ensure physical realism, we follow the logic of LI-GS and adopt 2D Gaussian Surfelszui as the foundational map representation \cite{ligs, 2dgs}. By explicitly constraining the normal direction, it produces better-fitting planar structures and suppresses floaters.

Each surfel $G_i$ is defined within its local tangent plane. For any point $(u, v)$ in the local 2D tangent space, the spatial distribution $d_{\sigma}$ follows $d_{\sigma}(u, v) = \exp \left( -\frac{u^2 + v^2}{2} \right)$.
The opacity contribution $\alpha_i$ of a surfel at a specific pixel is determined by the product of its intrinsic opacity $o_i$ and this spatial distribution $\alpha_i(u, v) = o_i \cdot d_{\sigma}(u, v)$.
This formulation mathematically restricts the Gaussian energy to a 2D manifold by omitting the component perpendicular to the plane, providing a precise basis for subsequent geometric supervision via point cloud priors.

The appearance $C(p)$ at pixel $p$ is obtained through $\alpha$-blending of overlapping surfels:
\begin{equation}
C(p) = \sum_{i \in \mathcal{N}} c_i \alpha_i \prod_{j=1}^{i-1} (1 - \alpha_j)
\end{equation}
where $\mathcal{N}$ denotes the ordered set of surfels that intersect the ray passing through pixel $p$, and $c_i$ is the high-dimensional feature or color associated with the $i$-th surfel.

\subsection{Geometry Reinforcement via SOGMM Modeling}
In large-scale panoptic reconstruction, optimization relying solely on visual gradients often struggles with geometric uncertainties caused by drastic variations in view distance or textureless regions. To address this, we introduce the Self-Organizing Gaussian Mixture Model (SOGMM) to integrate absolute scale information from 3D point clouds. SOGMM serves not only as a structural foundation for surfel initialization but also as a continuous geometric reference field throughout the optimization cycle, ensuring high physical realism.


\textbf{Probabilistic SOGMM Representation.}
We fit the scene point clouds into a set of anisotropic Gaussian components using a recursive splitting strategy. Unlike discrete point clouds, SOGMM transforms spatial distributions into a continuous probability density function defined as:
\begin{equation}
P(z' | \Theta) = \sum_{k=1}^K \hat{\pi}_k \mathcal{N}(z' | \hat{\mu}_k, \hat{\Sigma}_k)
\end{equation}
where $z' = [u, v, 0, g]^\top$ denotes a 4D augmented vector defined in the local tangent space. Here, $(u, v)$ represents the 2D local coordinates transformed from a world-space 3D point $p_i \in \mathbb{R}^3$, and $g$ is the grayscale intensity derived from image projections. We adopt this 4D SOGMM formulation to simultaneously capture the geometric structure and surface texture, providing a richer prior for surfel initialization. 

$\Theta = \{\hat{\pi}_k, \hat{\mu}_k, \hat{\Sigma}_k\}_{k=1}^K$ denotes the set of mixing weights, means, and covariances. $K$ is the total number of components adaptively determined by the self-organizing process. Each Gaussian node $\mathcal{M}_k$ represents a local planar prior. 

\textbf{Normal Constraint.}
To precisely capture surface characteristics, we perform eigenvalue decomposition on the covariance matrix $\hat{\Sigma}_k$, yielding eigenvalues $\lambda_0 \ge \lambda_1 \ge \lambda_2$ and their corresponding eigenvectors $\mathbf{v}_0, \mathbf{v}_1, \mathbf{v}_2$. 
Any world-space point $p_i$ within the neighborhood of $\mathcal{M}_k$ can be projected onto the local tangent plane. The relationship is formulated as $p_i = \hat{\mu}_k + u_i \mathbf{v}_0 + v_i \mathbf{v}_1 + w_i \mathbf{v}_2$
where $u_i, v_i, w_i$ represent the coordinates in the local coordinate system. Specifically, $\mathbf{v}_0$ and $\mathbf{v}_1$ span the principal directions of the local plane (associated with $u_i, v_i$), while the eigenvector $\mathbf{v}_2$ corresponding to the smallest eigenvalue is defined as the local reference normal $\hat{\mathbf{n}}_k$. In a perfectly planar structure, the vertical offset $w_i$ tends toward zero. This representation provides explicit directional constraints for the 2DGS representation.

SOGMM initializes the central positions and rotations of 2D Gaussians by sampling the distribution parameters. Furthermore, it establishes a criterion for geometric consistency in 3D space. To represent the deviation of a surfel $G_i$ from its geometric prior, we define the distance function from the surfel center $\mu_i$ to the local tangent plane of the SOGMM:
\begin{equation}
d(\mu_i, \mathcal{M}_k) = |(\mu_i - \hat{\mu}_k) \cdot \hat{\mathbf{n}}_k|
\end{equation}
This function measures the drift of the surfel along the normal direction. This anchoring mechanism effectively rectifies surface drifts caused by visual matching errors, maintaining rigorous topological consistency in complex, large-scale environments.

\section{End-to-End Panoptic Reconstruction}

\subsection{Query-guided Instance Segmentation}
\label{sec:query_instance}

\begin{figure}
    \centering
    \includegraphics[width=0.8\linewidth]{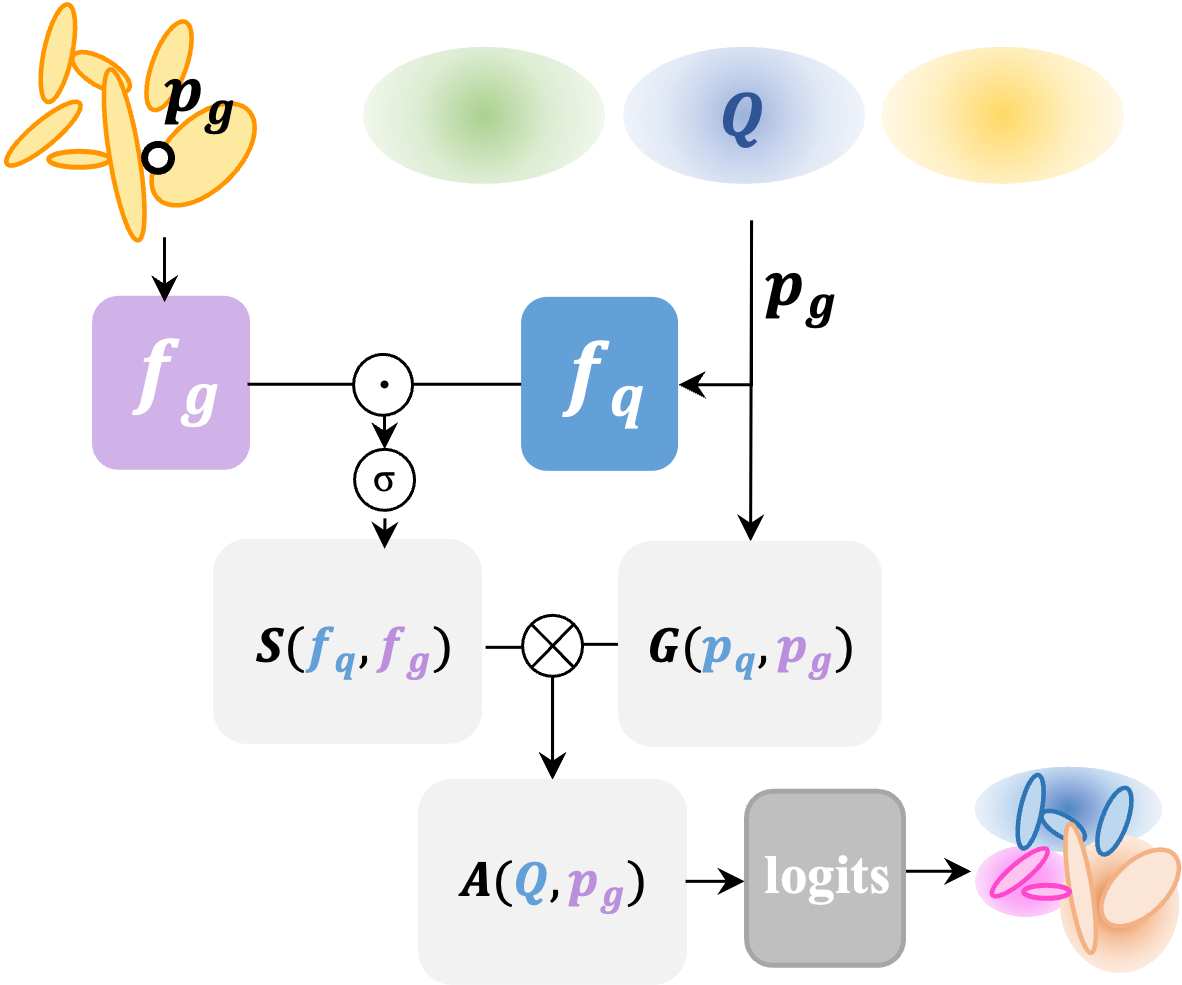}
    \caption{The instance branch utilizes a cross-attention mechanism between 3D Gaussian-modulated query tokens and SOGMM-reinforced 2DGS scene fields. }
    \label{fig:token}
\end{figure}

To bypass the error accumulation and heavy computational overhead inherent in traditional multi-stage pipelines, which rely on complex cross-frame data association, we propose an end-to-end architecture guided by explicit instance queries. By embedding semantic and instance-level information directly into the map representation, each 2D Gaussian surfel $G_i$ becomes a multi-modal primitive. Beyond traditional geometric attributes and appearance parameters, each surfel carries two additional learnable embeddings: a high-dimensional semantic feature vector $f_i^{sem}$ and an instance feature vector $f_i^{ins}$, forming a continuous neural feature field grounded on the scene geometry.

To extract discrete object instances from this field, we formulate the instance queries as explicitly localized representations. Following the paradigm of PanopticRecon++ \cite{pr++}, each instance query $\mathcal{Q}_k$ is defined as a dual-component entity: a high-dimensional query feature $f_q$ and a spatial 3D Gaussian distribution $\mathcal{G}_k(\mathbf{x}) = \mathcal{N}(\mathbf{x} | \mu_k, \Sigma_k)$, where $\mu_k$ and $\Sigma_k$ represent the learnable center and spatial covariance of the $k$-th instance.

To segment the scene Gaussians, we utilize a distance-aware cross-attention mechanism. Considering that the scene surfels are significantly smaller in scale than the Gaussian-modulated instance queries, we simplify each surfel $g$ into its center point $p_g$ during the interaction. The affinity between a query $q$ and a surfel $g$ is determined by fusing their feature similarity with their geometric correlation. First, the feature-based similarity is defined as:
\begin{equation}
    S(f_q, f_{g}) = \sigma(f_q^\top f_{g})
\end{equation}
where $\sigma(\cdot)$ denotes the sigmoid function. Simultaneously, as the instance queries follow a Gaussian distribution, the spatial relationship between a 3D point and the query can be described by the probability density. We define the geometric affinity between query $q$ and scene surfel $g$ as the ratio of their probability densities:
\begin{equation}
    D(p_q, p_g) = \frac{\varphi(p_g)}{\varphi(p_q)}
    \label{eq:weight}
\end{equation}
where $\varphi(\cdot)$ is the probability density function of the query's spatial Gaussian distribution, while $p_q$ and $p_g$ are the center coordinates of the instance query and the scene Gaussian, respectively. 

The resulting attention map $A(q, g)$ integrates these dimensions to ensure the assignment is both visually consistent and geometrically grounded:
\begin{equation}
    A(q, g) = S(f_q, f_{g}) D(p_q, p_g)
    \label{gsours}
\end{equation}
The discrete instance label for each surfel $g$ is then derived via a softmax operation over all $N$ queries:
\begin{equation}
    l_{ins}(g) = \text{Softmax}\left( [A(q_1, g), \dots, A(q_N, g)] \right)
\end{equation}
where $N$ is the total number of queries. Finally, to achieve consistent 2D panoptic segmentation, the pixel-level instance identity $I$ is rendered by $\alpha$-blending the labels of overlapping surfels along the camera ray:
\begin{equation}
    I = \sum_{i \in \mathcal{N}} l_{ins}(i) \alpha_i' \prod_{j=1}^{i-1} (1 - \alpha_j')
    \label{eq:ins_blending}
\end{equation}
This query-based grouping facilitates global panoptic consistency across the entire map without the need for explicit temporal tracking or heuristic data association.


\textbf{Frustum-constrained Cross-Attention.}
In large-scale environments, the dense population of 2D Gaussians makes global cross-attention computationally prohibitive. As the scene scale and the number of primitives grow, the attention matrix size leads to a memory explosion. To address this, we propose a local cross-attention mechanism constrained within the camera's view frustum $\mathcal{V}$, which restricts interactions to only relevant primitives during each training step.

Following the geometric criteria in 3DGS, we identify the active set of Gaussians within $\mathcal{V}$ as those whose $99\%$ confidence intervals intersect the current view frustum. For these selected primitives, we fuse the surfel central coordinates $\mu_j$, embedded via a positional encoding function $\psi(\cdot)$, with the instance features $f_j$ to construct the keys $K_j$ and values $V_j$:
\begin{equation}
    K_j = f_j + \psi(\mu_j), \quad V_j = f_j
\end{equation}
The attention score $e_{i,j}$ between a query $Q_i$ and a local surfel feature is then computed through a similarity projection:
\begin{equation}
    e_{i,j} = \frac{(Q_i W_Q)(K_j W_K)^\top}{\sqrt{d}}, \quad \forall j \in \mathcal{V}
\end{equation}
where $W_Q, W_K \in \mathbb{R}^{C \times d}$ are learnable projection matrices and $d$ is the feature dimension. The final query representation is updated by aggregating local surfel information:
\begin{equation}
    Q_i' = \text{Softmax}(\{e_{i,j}\}_{j \in \mathcal{V}}) \cdot (V_j W_V)
\end{equation}

To ensure high-performance execution, we implement this mechanism using a custom CUDA kernel. Upon filtering the valid Gaussians, we dispatch parallel threads for each primitive to compute the cross-attention scores and update the instance features. This localized strategy drastically reduces the peak training memory and shortens convergence time, particularly in expansive scenes. Following the attention computation, we leverage tile-based rasterization as defined in Eq. \ref{eq:ins_blending} to render the 2D instance maps. This anchoring of attention within the view frustum makes our model highly scalable for complex, high-resolution large-scale reconstructions.

\textbf{Dynamic Query Adjustment Strategy.} 
As the number of objects increases with spatial expansion in large-scale scenes, a fixed number of queries fails to balance computational efficiency with coverage completeness. We introduce a dynamic query adjustment scheme that adaptively prunes redundant instance tokens based on their utilization and geometric overlap. Specifically, we categorize redundant tokens into two types: useless and duplicate.

Since our initial instance masks are derived from potentially noisy open-vocabulary segmentation rather than ground-truth annotations, some queries may be initialized by erroneous masks. We define useless tokens as those that are never assigned or assigned only infrequently during the training process. By identifying and removing these low-utilization queries, we effectively reduce computational overhead while mitigating the propagation of segmentation noise.

In the early stages of optimization, multiple tokens may represent different portions of the same physical object due to the limited Field of View (FoV) of 2D images. As the representation converges, one query typically evolves to encompass the entire instance, rendering others redundant. To identify these duplicates, we introduce a 3D Intersection over Mask (IoM) metric. For a pair of tokens $Q_i$ and $Q_j$ with their respective 3D masks $M_i$ and $M_j$, the IoM is defined as:
\begin{equation}
    \text{IoM}_{ij} = \frac{M_i \cap M_j}{M_i}
\end{equation}
Our observations indicate that duplicate tokens often exhibit an enclosing behavior, where a larger 3D mask significantly overlaps with multiple smaller ones. By applying a threshold to the IoM, we retain only the query with the largest mask coverage, ensuring a concise and non-redundant instance representation.

\subsection{End-to-End Joint Optimization}

To achieve synergistic improvements in both scene geometry and panoptic understanding, we propose a multi-task joint optimization framework. By unifying geometric prior constraints and panoptic perceptual supervision, our system refines the spatial parameters of surfels and high-dimensional feature fields simultaneously through end-to-end backpropagation.

\textbf{Geometric and Photometric Supervision.} 
To ensure visual realism, we supervise the rendered images using a composite photometric loss: $\mathcal{L}_{rgb} = (1-\lambda_{s})\mathcal{L}_1 + \lambda_{s}\mathcal{L}_{ssim}$. More importantly, we leverage the local planar priors provided by the SOGMM to enforce geometric consistency through $\mathcal{L}_{geo}$:
\begin{equation}
\mathcal{L}_{geo} = \sum_{i} d(\mu_i, \mathcal{M}_k) + \lambda_{n}(1 - |\mathbf{n}_i \cdot \hat{\mathbf{n}}_k|)
\end{equation}
where $d(\mu_i, \mathcal{M}_k)$ is the point-to-plane distance. This term anchors the surfel center $\mu_i$ to the physical surface while aligning the surfel normal $\mathbf{n}_i$ with the reference normal $\hat{\mathbf{n}}_k$. Additionally, to maintain the structural integrity and compactness of the 2D primitives, we introduce an anisotropy regularization $\mathcal{L}_{iso} = \sum \|s_{i,1} - s_{i,2}\|^2$, which penalizes excessive stretching by minimizing the difference between the principal scaling axes $(s_1, s_2)$ of each surfel.

\textbf{Panoptic Feature Supervision.} 
The training of the semantic and instance fields is achieved by comparing rendered outputs with 2D pseudo-labels. Since 2D instance masks lack inter-view correspondence, we utilize the Hungarian Algorithm for linear assignment on a per-frame basis to associate instance queries with pseudo-GT labels. Following optimal matching, the instance branch is supervised via a combination of Dice and Binary Cross-Entropy (BCE) losses, while the semantic field utilizes standard Cross-Entropy (CE) loss:
\begin{equation}
    \mathcal{L}_{ins} = \sum_{k}  \mathcal{L}_{dice}(M_k, M_k^{gt}) + \mathcal{L}_{bce}(M_k, M_k^{gt}) 
\end{equation}

\begin{equation}
    \mathcal{L}_{sem} = \mathcal{L}_{ce}(M_{sem}, M_{sem}^{gt})
\end{equation}

\textbf{Total Optimization Objective.} 
The final joint objective function $\mathcal{L}_{total}$ is a weighted sum of the components, facilitating a complementary gradient flow between geometry and perception:
\begin{equation}
\mathcal{L}_{total} = \lambda_1 \mathcal{L}_{rgb} + \lambda_2 \mathcal{L}_{geo} + \lambda_3 \mathcal{L}_{ins} + \lambda_4 \mathcal{L}_{sem} + \lambda_5 \mathcal{L}_{iso}
\end{equation}
This multi-task training regime allows the system to reconstruct geometrically precise surfaces while maintaining robust, query-consistent panoptic understanding in large-scale environments.

\section{Efficient Rendering for High-Dimensional Features}
In large-scale panoptic reconstruction, semantic and instance features often possess extremely high dimensions, imposing significant computational pressure and memory bandwidth bottlenecks on the traditional rasterization pipeline. To meet the real-time requirements of robotic control loops while ensuring high-fidelity understanding, we optimize the rendering logic for high-dimensional feature fields.

\textbf{Precise Tile Intersection.}
Traditional rasterization often employs isotropic bounding spheres to determine intersections between Gaussians and pixel tiles, which introduces significant computational redundancy for flattened 2D surfels. To optimize this, we derive a compact Axis-Aligned Bounding Box (AABB) directly from the 2D projected covariance matrix $\Sigma'$. Specifically, for a given confidence threshold $\chi^2$ (e.g., $3\sigma$), the influence range $(\Delta x, \Delta y)$ of a surfel is defined by the diagonal elements of the projected covariance:
\begin{equation}
\Delta x = \sqrt{\chi^2 \Sigma'_{11}}, \quad \Delta y = \sqrt{\chi^2 \Sigma'_{22}}
\end{equation}
This strategy precisely filters out surfels that do not overlap with the current tile in the projection plane. By substituting loose spheres with tight AABBs, we reduce invalid Gaussian assignments by approximately 40\%, significantly alleviating memory bandwidth pressure in expansive large-scale environments.

\textbf{Top-K Hard Selection Strategy.}
Unlike RGB rendering, which requires dense blending to capture high-frequency lighting, dense semantic prediction tasks are less sensitive to high-frequency noise and are grounded in parsimonious geometric surfaces. Given that the feature dimensionality $C$ is substantially higher than RGB, accumulating all $N$ overlapping Gaussians becomes a primary computational bottleneck. Capitalizing on the inherent geometric sparsity of 2DGS, we propose a Top-$K$ Hard Selection mechanism. For each pixel, we restrict feature accumulation to a fixed set $\mathcal{K}$ containing the $K$ Gaussians with the highest opacity weights $\omega_j = \alpha_j \prod_{k=1}^{j-1}(1-\alpha_k)$:
\begin{equation}
F_{pixel} = \sum_{j \in \mathcal{K}} f_j \cdot \omega_j, \quad \text{where } |\mathcal{K}| = K
\end{equation}
By replacing the $O(N)$ accumulation with a constant $O(K)$ selection, we fundamentally streamline the high-dimensional vector multiply-accumulate operations. 

\section{Experiments}
\label{sec:experiments}

\begin{figure*}
    \centering
    \includegraphics[width=\linewidth]{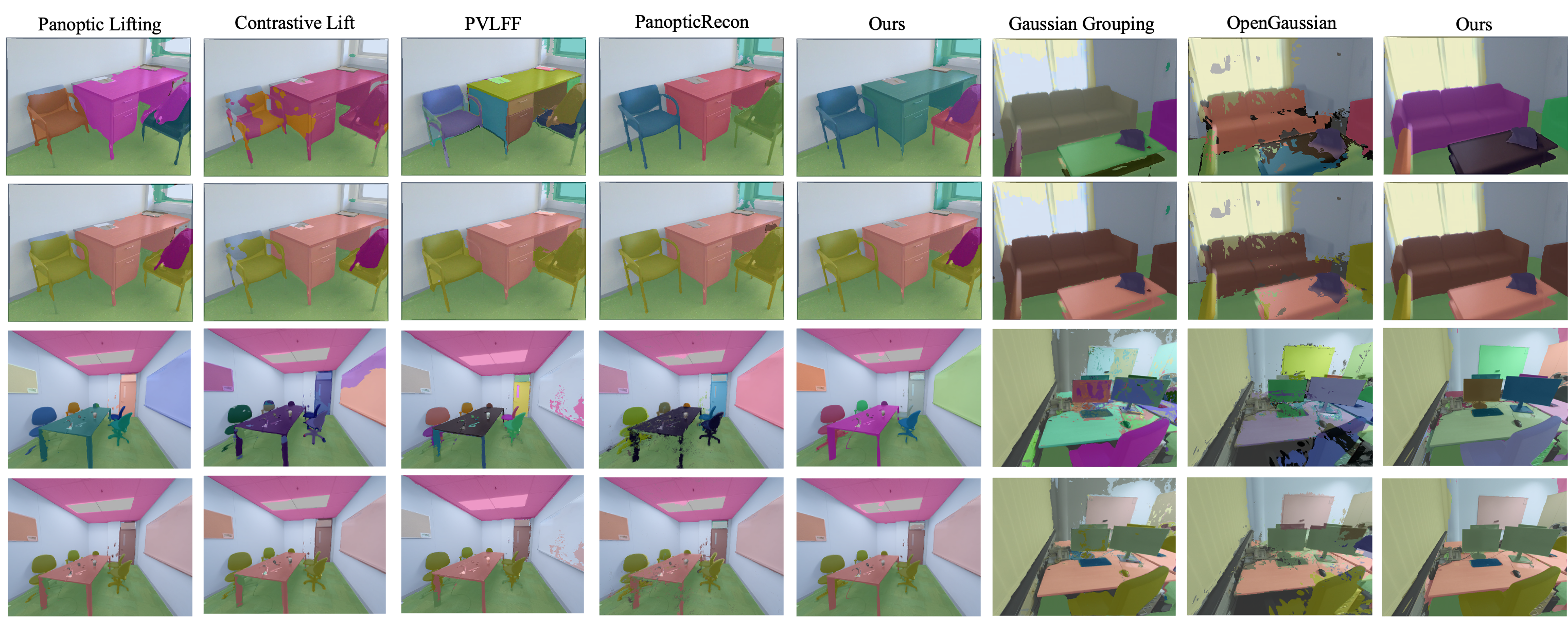}
    \caption{Comparison of the quality of semantic segmentation and panoptic segmentation of different methods on ScanNet V2 and ScanNet++. }
\end{figure*}

\subsection{Experimental Setup}

To validate the effectiveness of our proposed framework, we conduct extensive experiments focusing on geometric reconstruction accuracy, panoptic segmentation quality, and real-time rendering performance. We compare our system against several state-of-the-art methods in both outdoor and indoor environments to demonstrate its robustness and scalability.

To verify the efficacy of $\Psi$-Map across diverse environments, we conduct evaluations on three authoritative real-world benchmarks:
We utilize the KITTI-360 \cite{liao2022kitti} dataset, which comprises extensive mobile LiDAR scans and synchronized fisheye images from urban street scenes. This dataset is instrumental for assessing our SOGMM-guided geometric initialization and the robustness of our query-based panoptic mapping in expansive outdoor settings.
For fine-grained reconstruction and dense prediction, we employ ScanNet V2 \cite{dai2017scannet} and its higher-fidelity successor, ScanNet++ \cite{yeshwanth2023scannet++}. ScanNet V2 offers a vast array of RGB-D sequences with comprehensive semantic labels, while ScanNet++ provides sub-millimeter geometry and high-resolution textures. These datasets serve as the primary testbed for evaluating our Top-$K$ rendering efficiency and surface alignment accuracy. 

We employ a comprehensive set of metrics to evaluate different aspects of the system. For geometric reconstruction, we measure the Accuracy, Completeness, and F-score to assess how well the 2D surfels represent the physical surfaces. For panoptic understanding, we adopt standard metrics including Mean Intersection-over-Union (mIoU) for semantic segmentation, and Panoptic Quality (PQ), Segmentation Quality (SQ), and Recognition Quality (RQ) for instance-level evaluation. Additionally, we analyze the system efficiency using Frames Per Second (FPS) and GPU memory consumption to highlight our real-time performance.
All experiments and performance benchmarks are conducted on a workstation equipped with an NVIDIA RTX A6000 GPU.

\subsection{Geometric Reconstruction Evaluation}

To assess the geometric fidelity of $\Psi$-Map, we compare it against four LiDAR-only pipelines and nine state-of-the-art Gaussian-based methods. For a fair comparison, all Gaussian baselines are initialized with colorized LiDAR point clouds and augmented with sky masks for outdoor scenes.

As shown in Tab.~\ref{tab:geometry_results}, $\Psi$-Map outperforms all baselines in Accuracy, Completeness, and F-score. Qualitative results indicate that while LiDAR-only methods are limited by point sparsity, $\Psi$-Map utilizes visual information to reconstruct denser surfaces. Furthermore, our SOGMM-guided normalization mitigates the geometric degradation caused by excessive photometric loss in traditional Gaussian methods, effectively reducing artifacts and floaters.

Compared to standard 2DGS, which tends to oversmooth large-scale details, our framework preserves fine-grained structures through SOGMM's local planar priors. $\Psi$-Map maintains competitive rendering quality with a minor computational trade-off in optimization time due to the nearest neighbor search, which is justified by the significant gains in surface accuracy and structural consistency.

\begin{table}[htbp]
\centering
\caption{Geometric quality comparison on self-collected datasets.}
\label{tab:geometry_results}
\resizebox{\columnwidth}{!}{
\begin{tabular}{@{}l|ccc|ccc@{}}
\toprule
\textbf{Method} & \textbf{Acc. $\downarrow$} & \textbf{Comp. $\downarrow$} & \textbf{C-L1 $\downarrow$} & \textbf{Recall $\uparrow$} & \textbf{Prec. $\uparrow$} & \textbf{F1 $\uparrow$} \\ \midrule
2DGS \cite{2dgs} & 13.98 & 18.60 & 16.28 & 71.65 & 80.85 & 75.82 \\
Gaussian Surfels \cite{gaussiansurfels} & 14.25 & 35.95 & 25.10 & 70.47 & 48.33 & 57.29 \\
PGSR \cite{chen2024pgsr} & 13.52 & 19.63 & 16.58 & 72.66 & 73.67 & 73.08 \\
RaDe-GS \cite{radegs} & 13.28 & 17.13 & 15.22 & 73.42 & 78.59 & 75.88 \\
LIV-GaussMap \cite{hong2024liv} & 14.17 & 15.48 & 14.80 & 72.54 & 80.33 & 76.18 \\
GOF \cite{gof} & 14.27 & 15.37 & 14.83 & 70.55 & 80.20 & 75.04 \\
SuGaR \cite{guedon2024sugar} & 12.88 & 15.62 & 14.23 & 75.02 & 81.86 & 78.23 \\
Trim2DGS \cite{fan2024trim} & 14.65 & 33.75 & 25.30 & 68.64 & 61.12 & 64.09 \\
Trim3DGS \cite{fan2024trim} & 14.47 & 16.28 & 15.37 & 71.09 & 77.52 & 74.08 \\ \midrule
Ours  & 4.37 & 4.63 & 4.51 & 97.49 & 98.65 & 98.07 \\  \bottomrule
\end{tabular}
}
\end{table}

\begin{figure}
    \centering
    \includegraphics[width=\linewidth]{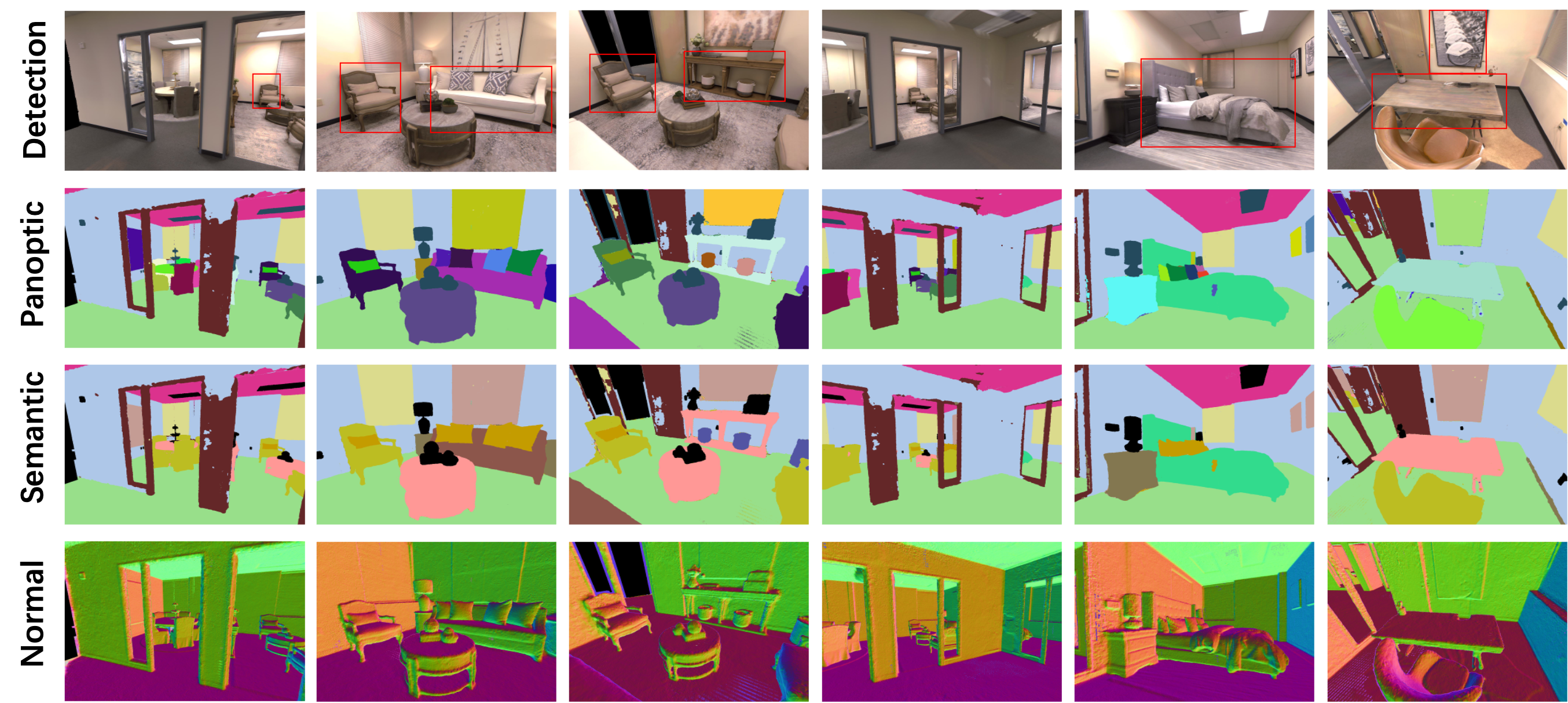}
    \caption{Qualitative rendering results on the Replica dataset, demonstrating $\Psi$-Map's ability to produce high-fidelity geometry, semantic, and panoptic maps. }
\end{figure}

\subsection{Panoptic Segmentation Performance}
To evaluate the scene understanding capabilities of $\Psi$-Map, we conduct a quantitative comparison against both NeRF-based lifting methods and Gaussian-based architectures. The results on ScanNet-V2 and ScanNet++ benchmarks are summarized in Tab.~\ref{tab:Comparative_baseline}.

As indicated by the metrics, $\Psi$-Map significantly outperforms all baselines across both datasets. In ScanNet-V2 and ScanNet++, our method achieves a state-of-the-art Panoptic Quality. The performance gain can be attributed to several key architectural advantages:

\textbf{End-to-End Joint Optimization.} Traditional multi-stage pipelines, such as PanopticRecon and Gaussian Grouping, often suffer from cascaded errors where inaccuracies in initial 2D segmentation masks propagate through the 3D reconstruction process. In contrast, $\Psi$-Map employs a unified joint optimization framework. By simultaneously refining geometry and high-dimensional panoptic features, our model allows perception and spatial structure to supervise each other, leading to a more globally consistent and noise-robust representation.

\textbf{Query-Based Temporal Consistency.} A major challenge for baselines like Panoptic Lifting and OpenGaussian is maintaining persistent instance identities across diverse viewpoints, frequently leading to identity switches or fragmentation. $\Psi$-Map resolves this by utilizing learnable 3D instance queries. This query-guided mechanism anchors each physical object to a unique identity that integrates both appearance and spatial priors. Consequently, our method ensures that multi-view observations of the same object are fused into a coherent 3D instance, as reflected by our nearly perfect RQ in ScanNet-V2.

\textbf{Distance-Aware Spatial Reasoning.} Feature-lifting approaches like Contrastive Lift often struggle to separate adjacent objects with similar visual textures. While OpenGaussian attempts to use spatial clustering, its priors are often too coarse for precise boundaries. Our framework leverages deformable instance queries and distance-aware similarity measures, enabling the system to perform high-resolution spatial reasoning. This allows $\Psi$-Map to effectively distinguish between spatially neighboring objects even under challenging photometric conditions.

\begin{table*}[t]
\centering
\caption{Quantitative comparison of panoptic segmentation performance on ScanNet-V2 and ScanNet++ datasets.}
\label{tab:Comparative_baseline}
\setlength{\tabcolsep}{3.5pt} 
\renewcommand\arraystretch{1.2}
\resizebox{\linewidth}{!}{
\begin{tabular}{l|ccccccc|ccccccc}
\toprule
\multirow{2}{*}{\textbf{Method}} & \multicolumn{7}{c|}{ScanNet-V2} & \multicolumn{7}{c}{ScanNet++} \\
\cmidrule(r){2-8} \cmidrule(l){9-15}
 & PQ $\uparrow$ & SQ $\uparrow$  & RQ $\uparrow$ & mIoU $\uparrow$ & mAcc $\uparrow$ & mCov $\uparrow$ & mW-Cov $\uparrow$ & PQ $\uparrow$ & SQ $\uparrow$ & RQ $\uparrow$ & mIoU $\uparrow$ & mAcc $\uparrow$ & mCov $\uparrow$ & mW-Cov $\uparrow$ \\
\midrule
\rowcolor[gray]{.95} \multicolumn{15}{l}{\textit{NeRF-based Lifting Methods}} \\
Panoptic Lifting \cite{lifting} & 57.86 & 61.96 & 85.31 & 67.91 & 78.59 & 45.88 & 59.93 & 71.14 & 77.48 & 88.14 & 81.34 & \textbf{89.67} & 56.17 & 68.51 \\
Contrastive Lift \cite{bhalgat2023contrastive} & 37.35 & 41.91 & 57.60 & 64.77 & 75.80 & 13.21 & 23.26 & 47.58 & 57.23 & 65.81 & 81.09 & 89.30 & 27.39 & 36.51 \\
PVLFF \cite{Chen2024PVLFF}                      & 30.11 & 51.71 & 44.43 & 55.41 & 63.96 & 45.75 & 48.41 & 52.24 & 66.86 & 65.56 & 62.53 & 70.31 & 67.95 & 75.47 \\
PanopticRecon \cite{yu2024panopticrecon}         & 63.70 & 64.81 & 81.17 & 68.62 & 80.87 & 66.58 & 77.84 & 68.29 & 77.01 & 85.05 & 77.75 & 87.08 & 51.34 & 62.79 \\
\midrule
\rowcolor[gray]{.95} \multicolumn{15}{l}{\textit{Gaussian-based Methods}} \\
Gaussian Grouping \cite{gaussiangrouping} & 43.75 & 50.63 & 72.68 & 58.05 & 68.68 & 52.70 & 58.10 & 33.10 & 40.60 & 67.27 & 59.53 & 68.13 & 29.83 & 36.83 \\
OpenGaussian \cite{opengaussian}           & 48.73 & 51.48 & 88.10 & 54.05 & 68.43 & 44.43 & 49.60 & 51.03 & 56.93 & 85.73 & 61.80 & 73.97 & 50.00 & 51.02 \\
\midrule
\textbf{$\Psi$-Map (Ours)} & \textbf{74.12} & \textbf{74.12} & \textbf{100.0} & \textbf{74.93} & \textbf{83.07} & \textbf{72.89} & \textbf{79.14} & \textbf{77.09} & \textbf{82.36} & \textbf{93.60} & \textbf{81.77} & 89.01 & \textbf{74.23} & \textbf{77.91} \\
\bottomrule
\end{tabular}
}
\end{table*}

\subsection{Rendering Efficiency and Real-time Inference}

In robotic simulation, the real-time synthesis of panoptic segmentation masks from a camera's perspective is vital for high-response, low-latency autonomous operations. Our experiments quantify how these optimizations collectively alleviate computational bottlenecks to ensure fluid interactive rates.

Table \ref{tab:ablation_speed} summarizes the performance across different configurations, tracking inference time (ms), frame rate (FPS), and efficiency metrics such as total rendered surfel count (RN-Total) and average surfels per tile (RN/Tile).

\begin{table}[t]
\centering
\caption{Study on Precise Tile Intersection and Top-$K$ Hard Selection.}
\label{tab:ablation_speed}
\resizebox{\columnwidth}{!}{
\begin{tabular}{lccccc}
\toprule
\textbf{Config} & \textbf{Time (ms)} $\downarrow$ & \textbf{FPS} $\uparrow$ & \textbf{RN-Total} $\downarrow$ & \textbf{RN / Tile} $\downarrow$ & \textbf{PQ} $\uparrow$ \\
\midrule
 Baseline & 46 & 22 & 6.1M & 1230 & 74.75 \\
 w/ Precise Tile & 40 & 25 & 3.9M & 800 & 74.73 \\
 w/ Top-$K$ & 27 & 37 & 1.2M & 610 & 74.17 \\
 Full Method & \textbf{22} & \textbf{45} & \textbf{1.2M} & \textbf{580} & \textbf{74.12} \\
\bottomrule
\end{tabular}
}
\end{table}

\textbf{Precise Tile Intersection.} The effectiveness of our enhanced intersection mechanism is demonstrated in the second row of Tab.~\ref{tab:ablation_speed}. We significantly filter out redundant tile-to-surfel assignments. Compared to the baseline setting, this mechanism reduces the total allocation count from 6.1M to 3.9M. This reduction proves that precise geometric intersection effectively mitigates the rasterization bottleneck inherent in standard 2DGS, enhancing overall throughput.

\textbf{Top-$K$ Hard Selection.} We further validate the strategy for tackling the feature accumulation bottleneck, which typically scales with the high channel count $C$ of panoptic fields. As shown in the \textit{w/ Top-K} configuration, we leverage the surface-aligned nature of 2DGS to truncate the blending process, selecting only the $K$ most relevant Gaussians per pixel. Comparing this setting to the baseline, the total rendered surfel count drops drastically to 1.2M, with inference latency falling to 27 ms, demonstrating that the Top-$K$ strategy provides substantial acceleration for high-dimensional feature rendering without a meaningful loss in PQ.

The last row of Tab.~\ref{tab:ablation_speed} integrates both optimizations, effectively addressing bottlenecks in both the tile assignment and alpha-blending stages. This combined approach achieves a low latency of 22 ms (45 FPS), meeting the rigorous real-time requirements for high-fidelity robotic navigation and simulation.


\begin{table}[ht]
\caption{Ablation study on different geometric representations on ScanNet-V2.}
\label{tab:Comparative_basemodel}
\centering
\setlength{\tabcolsep}{4pt}
\renewcommand\arraystretch{1.2}
\resizebox{\linewidth}{!}{
\begin{tabular}{l|ccc|cc|cc}
\toprule
\textbf{Base Model} & PQ $\uparrow$ & SQ $\uparrow$ & RQ $\uparrow$ & mIoU $\uparrow$ & mAcc $\uparrow$ & mCov $\uparrow$ & mW-Cov $\uparrow$ \\
\midrule
3DGS \cite{kerbl20233dgs} & 73.83 & 74.08 & 99.33 & \textbf{75.08} & 83.68 & 72.20 & 76.43 \\
2DGS \cite{2dgs} & 73.30 & 73.45 & 99.93 & 73.75 & 82.58 & 71.28 & 75.98 \\
Ours & \textbf{74.75} & \textbf{74.75} & \textbf{100.0} & 74.95 & \textbf{83.70} & \textbf{73.18} & \textbf{79.63} \\
\bottomrule
\end{tabular}
}
\end{table}

\subsection{Ablation Study}

In this section, we perform a series of ablation studies to evaluate the individual contributions of our key components. We analyze the impact of different geometric representations, token-based object modeling, spatial priors, and the dynamic adjustment mechanism to substantiate the design choices of the $\Psi$-Map framework.

\textbf{Impact of Geometric Representation.}
We investigate the influence of various geometric representations on final panoptic understanding performance. As shown in Tab.~\ref{tab:Comparative_basemodel}, while our query-guided architecture is compatible with different Gaussian models, the configuration utilizing our proposed geometric representation achieves the best results across all metrics, notably outperforming 2DGS by 1.45\% in PQ. This performance gain is primarily attributed to the high-fidelity surface reconstruction and the effective suppression of floaters through our SOGMM-guided constraints. By providing a cleaner and more physically accurate geometric backbone, our representation ensures that instance queries are anchored to precise structural surfaces rather than erroneous artifacts, thereby enabling more consistent instance association and semantic lifting. Consequently, we adopt this representation as the default geometric base for $\Psi$-Map.

\subsection{Case Study}
To demonstrate the practical utility of $\Psi$-Map, we evaluate its performance on a 3D Object Completion and Generation task using the Scan2CAD dataset, and on an Object Goal Navigation task in our self-collected and reconstructed indoor scenes.

\begin{table*}[]
\caption{Quantitative comparisons on Scan2CAD dataset}
\centering
\setlength{\tabcolsep}{7pt}
\renewcommand\arraystretch{1.2}
\begin{tabular}{lccccc|cc}
\toprule
Method & CD-S$\downarrow$ & F-Score-S$\uparrow$ & CD-O$\downarrow$ & F-Score-O$\uparrow$ & IoU-B$\uparrow$ & Col-O$\downarrow$ & Col-S$\downarrow$ \\
\midrule
AdaPoinTr
    & 0.018 & 0.841 & 0.062 & 0.532 & 0.597 & 0.329 & 0.023  \\
InstPIFu
    & 0.128 & 0.493 & 0.089 & 0.470 & 0.207 & 0.510 & 0.159  \\
Gen3DSR
    & 0.070 & 0.655 & 0.103 & 0.430 & 0.398 & 0.291 & 0.093  \\
Ours  
    & \textbf{0.016} & \textbf{0.903} & \textbf{0.036} & \textbf{0.710} & \textbf{0.666} & \textbf{0.203} & \textbf{0.015} \\
\bottomrule
\end{tabular}
\label{tab:Comparative_baseline}
\end{table*}

\begin{figure}
    \centering
    \includegraphics[width=\linewidth]{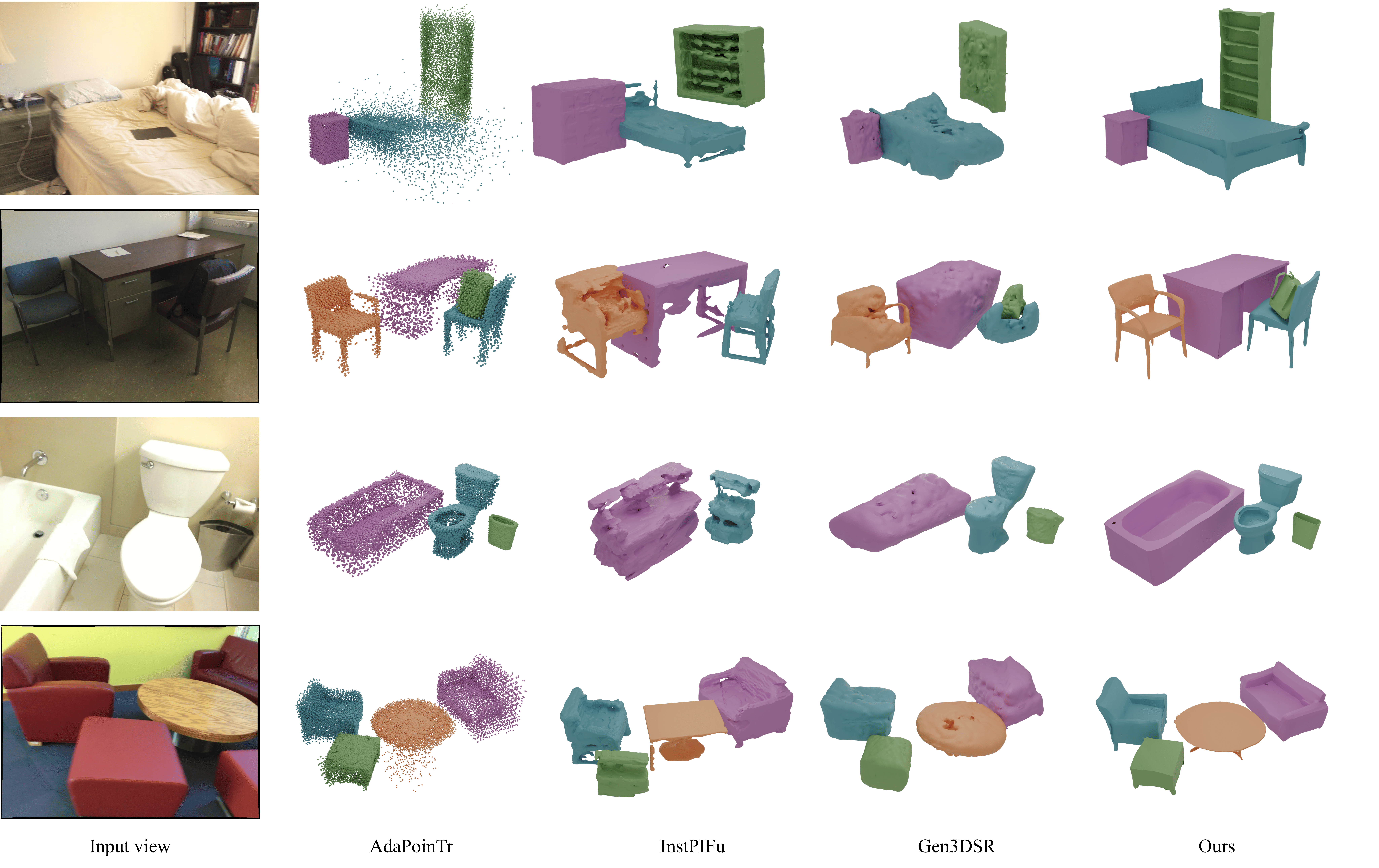}
    \caption{Quantitative comparisons on Scan2CAD dataset. }
    \label{fig:3Dobject}
\end{figure}

\textbf{3D Object Completion and Generation.} 
Our framework extends beyond segmentation reconstruction to support the generative completion of individual 3D objects within a scene following instance segmentation. A key challenge in this task is maintaining the fidelity of the observed geometry while inferring the complete shape of occluded or partially observed instances. As shown in Tab.~\ref{tab:Comparative_baseline}, our method significantly outperforms baseline approaches across all metrics. 

By leveraging the high-quality partial geometry from $\Psi$-Map as a structural prior, our generative pipeline ensures not only the geometric completeness of reconstructed objects but also their pose consistency relative to the original scene. Crucially, our approach explicitly considers the spatial relationship between the generated object and its surrounding environment, effectively avoiding geometric collisions or intersections with adjacent structures. 

Qualitative comparisons in Fig. \ref{fig:3Dobject} illustrate that, compared to existing methods, our solution produces complete 3D entities that are more consistent with the global scene context. Specifically, it excels at recovering the original orientations and relative poses of objects, ensuring that the completed models fit naturally into the environment without violating physical constraints. This capability makes $\Psi$-Map a powerful tool for high-fidelity simulation and digital twin synthesis.

\begin{figure*}
    \centering
    \includegraphics[width=\linewidth]{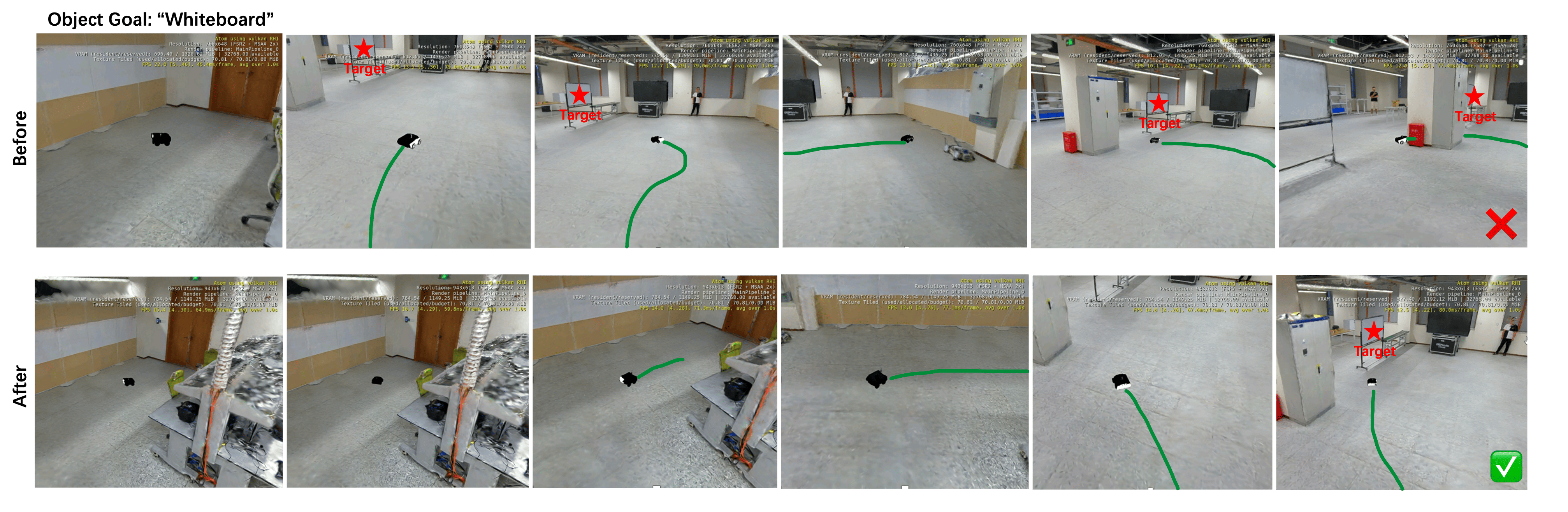}
    \caption{Comparison of navigation performance before and after fine-tuning with 3D-consistent labels.}
    \label{fig:nav}
\end{figure*}

\textbf{Object Goal Navigation.} 
We utilize a two-stage VLFM-based \cite{vlfm} pipeline, where the baseline perception module \cite{lei2025yolov13realtimeobjectdetection, sam} initially suffers from frequent missed or false detections due to limited single-frame visibility, resulting in low success rates as shown in Tab. \ref{tab:nav}.

To address this, we leverage $\Psi$-Map's 3D-consistent reconstruction as a high-fidelity ground truth generator. By projecting 3D-consistent labels back into 2D viewpoints, we create robust synthetic supervision that inherently benefits from multi-view fusion. Fine-tuning the perception module with this view-invariant knowledge significantly improves detection accuracy. As shown in Tab.~\ref{tab:nav}, this data-driven enhancement boosts the navigation success rate to 100\% in the Laboratory Room, proving that $\Psi$-Map can serve as a powerful engine to bolster robotic perception in complex deployments. 
As visualized in Fig. \ref{fig:nav}, after fine-tuning, the navigation algorithm benefits from significantly more accurate real-time visual perception. The agent is able to identify target objects more promptly and navigate to their immediate vicinity with higher precision, proving that $\Psi$-Map can serve as a powerful engine to bolster robotic perception in complex deployments.

\begin{table}[ht]
\centering
\caption{Impact of fine-tuning with 3D-consistent labels on navigation success rate (SR).}
\label{tab:nav}
\begin{tabular}{lcc}
\toprule
\textbf{Scene} & \textbf{SR-Before} & \textbf{SR-After} \\
\midrule
Laboratory Room & 20\% & \textbf{100\%} \\
Private House   & 25\% & \textbf{50\%} \\
\bottomrule
\end{tabular}
\end{table}

\section{Conclusion}
In this paper, we presented $\Psi$-Map, a geometry-incorporated panoptic splatting framework designed for real-time, large-scale scene understanding. By leveraging LiDAR-guided SOGMM modeling and 2D Gaussian surfels, our method achieves high-fidelity surface reconstruction with rigorous physical consistency. We introduced a query-guided end-to-end learning architecture that effectively overcomes error accumulation in traditional multi-stage pipelines, enabling cross-view consistent panoptic field learning. Furthermore, by incorporating Precise Tile Intersection and Top-$K$ Hard Selection, $\Psi$-Map successfully mitigates the computational bottlenecks of high-dimensional feature accumulation, ensuring real-time performance.

Experimental results on both indoor and outdoor benchmarks demonstrate that $\Psi$-Map significantly outperforms state-of-the-art methods in geometric accuracy and panoptic quality while maintaining a frame rate exceeding 45 FPS. Our case study further validates that $\Psi$-Map-generated 3D-consistent labels can serve as a powerful data engine to bolster the perception robustness of downstream robotic navigation tasks. Overall, $\Psi$-Map provides an efficient and robust foundation for future research in active robotic perception and high-fidelity simulation.
\bibliographystyle{IEEEtran}
\bibliography{IEEEabrv,bibliography}

\begin{thebibliography}{10}
\providecommand{\url}[1]{#1}
\csname url@samestyle\endcsname
\providecommand{\newblock}{\relax}
\providecommand{\bibinfo}[2]{#2}
\providecommand{\BIBentrySTDinterwordspacing}{\spaceskip=0pt\relax}
\providecommand{\BIBentryALTinterwordstretchfactor}{4}
\providecommand{\BIBentryALTinterwordspacing}{\spaceskip=\fontdimen2\font plus
\BIBentryALTinterwordstretchfactor\fontdimen3\font minus \fontdimen4\font\relax}
\providecommand{\BIBforeignlanguage}[2]{{%
\expandafter\ifx\csname l@#1\endcsname\relax
\typeout{** WARNING: IEEEtran.bst: No hyphenation pattern has been}%
\typeout{** loaded for the language `#1'. Using the pattern for}%
\typeout{** the default language instead.}%
\else
\language=\csname l@#1\endcsname
\fi
#2}}
\providecommand{\BIBdecl}{\relax}
\BIBdecl

\bibitem{tao2024maniskill3}
S.~Tao, F.~Xiang, A.~Shukla, Y.~Qin, X.~Hinrichsen, X.~Yuan, C.~Bao, X.~Lin, Y.~Liu, T.-k. Chan \emph{et~al.}, ``Maniskill3: Gpu parallelized robotics simulation and rendering for generalizable embodied ai,'' \emph{arXiv preprint arXiv:2410.00425}, 2024.

\bibitem{mittal2025isaac}
M.~Mittal, P.~Roth, J.~Tigue, A.~Richard, O.~Zhang, P.~Du, A.~Serrano-Munoz, X.~Yao, R.~Zurbr{\"u}gg, N.~Rudin \emph{et~al.}, ``Isaac lab: A gpu-accelerated simulation framework for multi-modal robot learning,'' \emph{arXiv preprint arXiv:2511.04831}, 2025.

\bibitem{zhang2024uni-navid}
J.~Zhang, K.~Wang, S.~Wang, M.~Li, H.~Liu, S.~Wei, Z.~Wang, Z.~Zhang, and H.~Wang, ``Uni-navid: A video-based vision-language-action model for unifying embodied navigation tasks,'' \emph{arXiv preprint arXiv:2412.06224}, 2024.

\bibitem{li2024evaluating}
X.~Li, K.~Hsu, J.~Gu, K.~Pertsch, O.~Mees, H.~R. Walke, C.~Fu, I.~Lunawat, I.~Sieh, S.~Kirmani \emph{et~al.}, ``Evaluating real-world robot manipulation policies in simulation,'' \emph{arXiv preprint arXiv:2405.05941}, 2024.

\bibitem{wu2025rlgsbridge}
Y.~Wu, L.~Pan, W.~Wu, G.~Wang, Y.~Miao, F.~Xu, and H.~Wang, ``Rl-gsbridge: 3d gaussian splatting based real2sim2real method for robotic manipulation learning,'' in \emph{2025 IEEE International Conference on Robotics and Automation (ICRA)}.\hskip 1em plus 0.5em minus 0.4em\relax IEEE, 2025, pp. 192--198.

\bibitem{qureshi2025splatsim}
M.~N. Qureshi, S.~Garg, F.~Yandun, D.~Held, G.~Kantor, and A.~Silwal, ``Splatsim: Zero-shot sim2real transfer of rgb manipulation policies using gaussian splatting,'' in \emph{2025 IEEE International Conference on Robotics and Automation (ICRA)}.\hskip 1em plus 0.5em minus 0.4em\relax IEEE, 2025, pp. 6502--6509.

\bibitem{li2025rose}
P.~Li, H.~Geng, J.~Crate, Y.~Han, J.~Zhang, F.~Wang, C.~T. Cheng, R.~Dong, Y.-J. Wang, H.~Lou \emph{et~al.}, ``Rose: Reconstructing objects, scenes, and trajectories from casual videos for robotic manipulation,'' in \emph{NeurIPS 2025 Workshop on Bridging Language, Agent, and World Models for Reasoning and Planning}.

\bibitem{han2025re3sim}
X.~Han, M.~Liu, Y.~Chen, J.~Yu, X.~Lyu, Y.~Tian, B.~Wang, W.~Zhang, and J.~Pang, ``Re{$^3$}sim: Generating high-fidelity simulation data via 3d-photorealistic real-to-sim for robotic manipulation,'' \emph{arXiv preprint arXiv:2502.08645}, 2025.

\bibitem{xia2024video2game}
H.~Xia, Z.-H. Lin, W.-C. Ma, and S.~Wang, ``Video2game: Real-time interactive realistic and browser-compatible environment from a single video,'' in \emph{Proceedings of the IEEE/CVF Conference on Computer Vision and Pattern Recognition}, 2024, pp. 4578--4588.

\bibitem{wu2023objectsdf++}
Q.~Wu, K.~Wang, K.~Li, J.~Zheng, and J.~Cai, ``Objectsdf++: Improved object-compositional neural implicit surfaces,'' in \emph{Proceedings of the IEEE/CVF International Conference on Computer Vision}, 2023, pp. 21\,764--21\,774.

\bibitem{wang2025vggt}
J.~Wang, M.~Chen, N.~Karaev, A.~Vedaldi, C.~Rupprecht, and D.~Novotny, ``Vggt: Visual geometry grounded transformer,'' in \emph{Proceedings of the Computer Vision and Pattern Recognition Conference}, 2025, pp. 5294--5306.

\bibitem{feature3dgs}
S.~Zhou, H.~Chang, S.~Jiang, Z.~Fan, D.~Xu, P.~Chari, S.~You, Z.~Wang, and A.~Kadambi, ``Feature 3dgs: Supercharging 3d gaussian splatting to enable distilled feature fields,'' in \emph{Proceedings of the IEEE/CVF Conference on Computer Vision and Pattern Recognition}, 2024, pp. 21\,676--21\,685.

\bibitem{yu2024mip}
Z.~Yu, A.~Chen, B.~Huang, T.~Sattler, and A.~Geiger, ``Mip-splatting: Alias-free 3d gaussian splatting,'' in \emph{Proceedings of the IEEE/CVF conference on computer vision and pattern recognition}, 2024, pp. 19\,447--19\,456.

\bibitem{guedon2024sugar}
A.~Gu{\'e}don and V.~Lepetit, ``Sugar: Surface-aligned gaussian splatting for efficient 3d mesh reconstruction and high-quality mesh rendering,'' in \emph{Proceedings of the IEEE/CVF Conference on Computer Vision and Pattern Recognition}, 2024, pp. 5354--5363.

\bibitem{2dgs}
B.~Huang, Z.~Yu, A.~Chen, A.~Geiger, and S.~Gao, ``2d gaussian splatting for geometrically accurate radiance fields,'' in \emph{ACM SIGGRAPH 2024 conference papers}, 2024, pp. 1--11.

\bibitem{chen2024pgsr}
D.~Chen, H.~Li, W.~Ye, Y.~Wang, W.~Xie, S.~Zhai, N.~Wang, H.~Liu, H.~Bao, and G.~Zhang, ``Pgsr: Planar-based gaussian splatting for efficient and high-fidelity surface reconstruction,'' \emph{IEEE Transactions on Visualization and Computer Graphics}, 2024.

\bibitem{gmm}
K.~Goel, N.~Michael, and W.~Tabib, ``Probabilistic point cloud modeling via self-organizing gaussian mixture models,'' \emph{IEEE Robotics and Automation Letters}, vol.~8, no.~5, pp. 2526--2533, 2023.

\bibitem{gmm2}
K.~Goel and W.~Tabib, ``Incremental multimodal surface mapping via self-organizing gaussian mixture models,'' \emph{IEEE Robotics and Automation Letters}, vol.~8, no.~12, pp. 8358--8365, 2023.

\bibitem{clip}
A.~Radford, J.~W. Kim, C.~Hallacy, A.~Ramesh, G.~Goh, S.~Agarwal, G.~Sastry, A.~Askell, P.~Mishkin, J.~Clark \emph{et~al.}, ``Learning transferable visual models from natural language supervision,'' in \emph{International conference on machine learning}.\hskip 1em plus 0.5em minus 0.4em\relax PMLR, 2021, pp. 8748--8763.

\bibitem{kerr2023lerf}
J.~Kerr, C.~M. Kim, K.~Goldberg, A.~Kanazawa, and M.~Tancik, ``Lerf: Language embedded radiance fields,'' in \emph{Proceedings of the IEEE/CVF International Conference on Computer Vision}, 2023, pp. 19\,729--19\,739.

\bibitem{langsplat}
M.~Qin, W.~Li, J.~Zhou, H.~Wang, and H.~Pfister, ``Langsplat: 3d language gaussian splatting,'' in \emph{Proceedings of the IEEE/CVF Conference on Computer Vision and Pattern Recognition}, 2024, pp. 20\,051--20\,060.

\bibitem{gaussiangrouping}
M.~Ye, M.~Danelljan, F.~Yu, and L.~Ke, ``Gaussian grouping: Segment and edit anything in 3d scenes,'' in \emph{European Conference on Computer Vision}.\hskip 1em plus 0.5em minus 0.4em\relax Springer, 2024, pp. 162--179.

\bibitem{sam}
A.~Kirillov, E.~Mintun, N.~Ravi, H.~Mao, C.~Rolland, L.~Gustafson, T.~Xiao, S.~Whitehead, A.~C. Berg, W.-Y. Lo \emph{et~al.}, ``Segment anything,'' in \emph{Proceedings of the IEEE/CVF International Conference on Computer Vision}, 2023, pp. 4015--4026.

\bibitem{mask2former}
B.~Cheng, I.~Misra, A.~G. Schwing, A.~Kirillov, and R.~Girdhar, ``Masked-attention mask transformer for universal image segmentation,'' in \emph{CVPR}, 2022.

\bibitem{lee2024compact}
J.~C. Lee, D.~Rho, X.~Sun, J.~H. Ko, and E.~Park, ``Compact 3d gaussian representation for radiance field,'' in \emph{Proceedings of the IEEE/CVF Conference on Computer Vision and Pattern Recognition}, 2024, pp. 21\,719--21\,728.

\bibitem{ligs}
C.~Jiang, R.~Gao, K.~Shao, Y.~Wang, R.~Xiong, and Y.~Zhang, ``Li-gs: Gaussian splatting with lidar incorporated for accurate large-scale reconstruction,'' \emph{IEEE Robotics and Automation Letters}, 2024.

\bibitem{pr++}
X.~Yu, Y.~Xie, Y.~Liu, H.~Lu, R.~Xiong, Y.~Liao, and Y.~Wang, ``Leverage cross-attention for end-to-end open-vocabulary panoptic reconstruction,'' \emph{arXiv preprint arXiv:2501.01119}, 2025.

\bibitem{liao2022kitti}
Y.~Liao, J.~Xie, and A.~Geiger, ``Kitti-360: A novel dataset and benchmarks for urban scene understanding in 2d and 3d,'' \emph{IEEE Transactions on Pattern Analysis and Machine Intelligence}, vol.~45, no.~3, pp. 3292--3310, 2022.

\bibitem{dai2017scannet}
A.~Dai, A.~X. Chang, M.~Savva, M.~Halber, T.~Funkhouser, and M.~Nie{\ss}ner, ``Scannet: Richly-annotated 3d reconstructions of indoor scenes,'' in \emph{Proceedings of the IEEE conference on computer vision and pattern recognition}, 2017, pp. 5828--5839.

\bibitem{yeshwanth2023scannet++}
C.~Yeshwanth, Y.-C. Liu, M.~Nie{\ss}ner, and A.~Dai, ``Scannet++: A high-fidelity dataset of 3d indoor scenes,'' in \emph{Proceedings of the IEEE/CVF International Conference on Computer Vision}, 2023, pp. 12--22.

\bibitem{gaussiansurfels}
P.~Dai, J.~Xu, W.~Xie, X.~Liu, H.~Wang, and W.~Xu, ``High-quality surface reconstruction using gaussian surfels,'' in \emph{ACM SIGGRAPH 2024 conference papers}, 2024, pp. 1--11.

\bibitem{radegs}
B.~Zhang, C.~Fang, R.~Shrestha, Y.~Liang, X.~Long, and P.~Tan, ``Rade-gs: Rasterizing depth in gaussian splatting,'' \emph{arXiv preprint arXiv:2406.01467}, 2024.

\bibitem{hong2024liv}
S.~Hong, J.~He, X.~Zheng, and C.~Zheng, ``Liv-gaussmap: Lidar-inertial-visual fusion for real-time 3d radiance field map rendering,'' \emph{IEEE Robotics and Automation Letters}, vol.~9, no.~11, pp. 9765--9772, 2024.

\bibitem{gof}
Z.~Yu, T.~Sattler, and A.~Geiger, ``Gaussian opacity fields: Efficient adaptive surface reconstruction in unbounded scenes,'' \emph{ACM Transactions on Graphics (ToG)}, vol.~43, no.~6, pp. 1--13, 2024.

\bibitem{fan2024trim}
L.~Fan, Y.~Yang, M.~Li, H.~Li, and Z.~Zhang, ``Trim 3d gaussian splatting for accurate geometry representation,'' \emph{arXiv preprint arXiv:2406.07499}, 2024.

\bibitem{lifting}
Y.~Siddiqui, L.~Porzi, S.~R. Bul\`o, N.~M\"uller, M.~Nie{\ss}ner, A.~Dai, and P.~Kontschieder, ``Panoptic lifting for 3d scene understanding with neural fields,'' in \emph{Proceedings of the IEEE/CVF Conference on Computer Vision and Pattern Recognition (CVPR)}, June 2023.

\bibitem{bhalgat2023contrastive}
Y.~Bhalgat, I.~Laina, J.~F. Henriques, A.~Zisserman, and A.~Vedaldi, ``Contrastive lift: 3d object instance segmentation by slow-fast contrastive fusion,'' in \emph{NeurIPS}, 2023.

\bibitem{Chen2024PVLFF}
H.~Chen, K.~Blomqvist, F.~Milano, and R.~Siegwart, ``Panoptic vision-language feature fields,'' \emph{IEEE Robotics and Automation Letters (RA-L)}, vol.~9, no.~3, pp. 2144--2151, 2024.

\bibitem{yu2024panopticrecon}
X.~Yu, Y.~Liu, C.~Han, S.~Mao, S.~Zhou, R.~Xiong, Y.~Liao, and Y.~Wang, ``Panopticrecon: Leverage open-vocabulary instance segmentation for zero-shot panoptic reconstruction,'' \emph{arXiv preprint arXiv:2407.01349}, 2024.

\bibitem{opengaussian}
Y.~Wu, J.~Meng, H.~Li, C.~Wu, Y.~Shi, X.~Cheng, C.~Zhao, H.~Feng, E.~Ding, J.~Wang \emph{et~al.}, ``Opengaussian: Towards point-level 3d gaussian-based open vocabulary understanding,'' \emph{arXiv preprint arXiv:2406.02058}, 2024.

\bibitem{kerbl20233dgs}
B.~Kerbl, G.~Kopanas, T.~Leimk{\"u}hler, and G.~Drettakis, ``3d gaussian splatting for real-time radiance field rendering.'' \emph{ACM Trans. Graph.}, vol.~42, no.~4, pp. 139--1, 2023.

\bibitem{vlfm}
\BIBentryALTinterwordspacing
N.~Yokoyama, S.~Ha, D.~Batra, J.~Wang, and B.~Bucher, ``Vlfm: Vision-language frontier maps for zero-shot semantic navigation,'' 2023. [Online]. Available: \url{https://arxiv.org/abs/2312.03275}
\BIBentrySTDinterwordspacing

\bibitem{lei2025yolov13realtimeobjectdetection}
\BIBentryALTinterwordspacing
M.~Lei, S.~Li, Y.~Wu, H.~Hu, Y.~Zhou, X.~Zheng, G.~Ding, S.~Du, Z.~Wu, and Y.~Gao, ``Yolov13: Real-time object detection with hypergraph-enhanced adaptive visual perception,'' 2025. [Online]. Available: \url{https://arxiv.org/abs/2506.17733}
\BIBentrySTDinterwordspacing

\end{thebibliography}

\end{document}